# Winograd Schema - Knowledge Extraction Using Narrative Chains

Vatsal Mahajan
*Department of Computer Science*
*Arizona State University*
vmahaja1@asu.edu

**Abstract**— The Winograd Schema Challenge (WSC) is a test of machine intelligence, designed to be an improvement on the Turing test. A Winograd Schema consists of a sentence and a corresponding question. To successfully answer these questions, one requires the use of commonsense knowledge and reasoning. This work focuses on extracting common sense knowledge which can be used to generate answers for the Winograd schema challenge. Common sense knowledge is extracted based on events (or actions) and their participants; called Event-Based Conditional Commonsense (ECC). I propose an approach using Narrative Event Chains [Chambers *et al.*, 2008] to extract ECC knowledge. These are stored in templates, to be later used for answering the WSC questions. This approach works well with respect to a subset of WSC tasks.

—————————— ◆ ——————————

## 1 INTRODUCTION

THE Winograd Schema Challenge (WSC) poses a set of multiple-choice questions that have a particular form for example:

*Sentence: The trophy would not fit in the brown suitcase because it was too big (small).*
*Question: What was too big (small)?*
*Answer0: the trophy*
*Answer1: the suitcase*

To answer the above question, one requires to have the knowledge that an object being big would have a higher chance of not fitting a in suitcase, as compared to a small object. Here some external knowledge is required to help with this spatial reasoning. The primary focus is to extract common sense knowledge based on events (or actions) and their participants; called Event-Based Conditional Commonsense (ECC) [Sharma *et al.*, 2016]. Extracted knowledge is stored in the following format:

*X.**PROP** = true/false, may cause execution of **A** [ARG\*: X; ARG\*: Y]*

where PROP denotes the property causing the action *A*. *ARG\*: X* and *ARG\*: Y* denote the agent and recipient for the action *A*.

For example, consider the following sentence: *"Jim yelled at Kevin because Kevin was so upset"*. Where the event/action is *"yelled"* and the property is *"upset"* with the agent as *"Jim"* and recipient as *"Kevin"*. The extracted knowledge is stored as the following template:

*Jim.**upset** = true, may cause execution of **yelled** [ARG0: Jim; ARG1: Kevin]*

This knowledge can be used to resolve ambiguity in a task like:

*Sentence: Jim yelled at Kevin because he was so upset.*
*Question: Who was upset?*
*Answer0: Jim*
*Answer1: Kevin*

by concluding that *"he"* refers to *"Kevin"*.

Narrative Event Chains [Chambers *et al.*, 2008] are used to extract ECC from documents in the corpus. Narrative chains are partially ordered sets of events centered around a common protagonist. For example, consider a sequence of sentences as follows:

*"Kevin wanted the ball. Kevin gets the ball from John."*

where the common protagonist is *"Kevin"* and the events are *"wanted"* and *"gets"*. In above example *"wanted"* event is causing *"gets"* event. So, the causal knowledge is extracted from sequence of sentences as

*"Kevin.**wanted** = true, may cause execution of **gets** [ARG0: Kevin, ARG1: Ball]"*
Another common protagonist in this sentence is 'ball'.

This approach extracts a set of event pairs that share a common protagonist. Then labeled Timebank Corpus is used to create a supervised learning method to classify temporal relation between two events as before or after. Using this model, the unordered event set is ordered into a narration. Rather than creating a chain I simply extract the causal relations to create knowledge templates. The result of the extraction is used to answer WSC questions.

## 2 RELATED WORK

There are various approaches for recognizing causal relations which can be used to extract common sense knowledge. One approach recognizes these causal relations by using framenets [Aharon *et al.*, 2010]. FrameNet is a manually constructed database based on Frame Semantics. It models the semantic argument structure of predicates in terms of prototypical situations called frames. This approach utilizes FrameNet's annotated sentences and re-lations between frames to extract both the entailment relations and their argument mappings.

Another approach that uses event based commonsense knowledge extraction is "Automatic Extraction of Events-Based Conditional Commonsense Knowledge" [Sharma et al., 2016]. It takes OANC corpus and performs semantic parsing on sentences to extract entities, events and their causal relations. Then uses Answer set programming to represent common sense knowledge.

## 3 APPROACH

This approach extracts knowledge of the form "A.x causes B.y", where x and y are events that share a participant and A & B are actors. I assume that although a narrative has several participants, there is a central actor who characterizes a narrative chain: the protagonist. For example, *"The policeman searched the suspect and then arrested the suspect"*. In this example, there are two actors: *policeman* and *suspect*, and there are two events: *search* and *arrest*, where *search* and *arrest* share the same participant *policeman*. This sentence can be used to generate knowledge as follows:

*suspect.search = true, may cause execution of **arrest** (policeman, suspect)*

The system developed in this paper creates a chain of two events with a common protagonist and later uses this chain to create knowledge base using the template format described above. Learning these prototypical schematic sequences of events is important for rich understanding of text.

### 3.1 Events and their Participants

To generate a set of events and their participants, I extract a series of event tuples from the corpus. A tuple is of the form:

< (subject1 event1 object1), (subject2 event2 object2) >

where event1 and event2 share a common protagonist and subjectX, objectX refer to the participants of the eventX. After the completion of this step, a set of these tuples is generated. An important point to consider is that, the events in a tuple may or may not be ordered. The ordering is performed in the next step; temporal ordering is discussed in details in the section 3.3.

The corpus used here is the English Gigaword corpus, LDC Catalog No. LDC2003T05. The corpus file is an XML file with multiple <DOC> elements. Each <DOC> element represents a document and content of this document is encapsulated inside a <TEXT> element. Each of these text elements have multiple paragraphs contained inside <P> tags. The foremost task was to extract documents and sentences for each document.

Once the sentences are extracted, the next step is to extract a set of event tuples for each document, finding events across sentences in the documents. In this step, I find the relationship between different predicates or events in subsequent sentences of a paragraph. An event tuple is extracted in the following form:

<Subject, Verb, Object, Typed_dependency>

where verb represents the event and the typed_dependency can take two values {subject, object}. The typed dependency is a way to represent the propagonist. For two events, the protagonist can be a subject for an event and an object for another. The system also collects the pair of verbs/events which are connected through the same co-referring entity. This information is used to build a verb/dependency graph between various events and calculate how pairs of events are occurring together. Subject and Object are the actors involved in this event. The system only extracts knowledge where exactly two actors are involved in each event. Given this graph, verbs can be clustered to create narrative chains with multiple narrative events. For a document, the following steps are used:

1. Create Dependency Graph and generate POS: Sentences are parsed using the Stanford parser and the dependency graph is generated. Items will be punctuated as sentences where it is appropriate.
2. Co-reference Resolution: The system uses the openNLP libraries to find how many entities exist and to get the co-referring entities. After resolving coreference, a data structure is maintained to store the entity and the co-referring entities together.
3. Storing Event-Dependency Pair: In this step, the dependency graph is used to get events which are relating the co-referring entities. For each verb/event the type of relationship with entities (either "nsubj" or "dobj") is stored, which would be used for creating knowledge later.
4. Verb/dependency graph: In this step, the system creates a verb/dependency graph from the information collected above which can be used for pointwise mutual information(pmi) calculation. The value of pmi is calculated using the following formulae:

$$pmi\big(e(w,d), e(v,g)\big) = \log \big(\frac{P(e(w,d), e(v,g))}{P\big(e(w,d)\big).P\big(e(v,g)\big)}\big)$$

where e(w, d) is the verb/dependency pair w and d (hit/subj).

$$P\big(e(w,d), e(v,g)\big) = \frac{C(e(w,d), e(v,g))}{\sum_{x,y} \sum_{d,f} C(e(x,d), e(y,f))}$$

where C(e(x,d),e(y,f)) is the number of times the two events e(x,d) and e(y,f) had a co-referring entity with typed dependencies d and f. In the verb/dependency graph, each independent <event, typed dependency> tuple represents a node. Two nodes are connected if they have a common protagonist and the edge cost is the pmi score calculated per the above formula.

### 3.2 Event Chain

Currently, the system creates a graph which has all the detected verbs as nodes. A verb could have occurred multiple times with different typed dependencies (Subject or Object). Here event pairs with shared referring entity are used to create the knowledge base. Fig. 1 shows a sample unordered event chain with 2 events in it.

The output of this step creates an array of unordered event chains. The next section focuses on ordering these narrative chains.

```
"eventChain": [
    {
        "object": "ball",
        "verb": "throw",
        "subject": "john"
    },
    {
        "object": "william",
        "verb": "hit",
        "subject": "ball"
    }
]
```

Fig. 1. Sample unordered event chain with 2 events.

### 3.3 Temporal Relations

Here I will discuss the temporal classification of verb/event pairs. The Timebank Corpus labels events and binary relations between events representing temporal order. I used classifiers that follow standard feature-based machine learning approaches as described in [Mani et al., 2006; Chambers et al., 2007] with training data from Timebank Corpus.

ClearTK (a framework for developing machine learning and natural language processing) was used to get the temporal relations between events. The algorithm is described below:

1. Stage1: Learning Event Attributes: The system learns temporal attributes for events in the NYT Corpus using the labeled Timebank Corpus as training-data. Here it learns the five temporal attributes associated with these events as tagged in the Timebank Corpus. 1) Tense and; 2) grammatical aspect are necessary in any approach to temporal ordering as they Timebank Corpus define both temporal location and structure of the event; 3) event class is the type of event. Table 1 lists the features used [Chambers et al., 2007]. Naive Bayes with Laplace smoothing is used to predict the value all 3 attributes. Three classifiers were used one for each of the attributes.

2. Stage2: Learning Event-Event Features: Here the system learns the temporal relation between events (before/after). Again, the TimeBank Corpus is used as training-data. The features [Chambers et al., 2007] used are: 1) Event Specific: The five temporal attributes from Stage 1 are used for each event in the pair, as well as the event strings, lemmas, and WordNet synsets; 2) POS tags; 3) Event-Event Syntactic Properties: A phrase P is said to dominate another phrase Q if Q is a daughter node of P in the syntactic parse tree. Dominance is taken as a feature which takes values on/off; 4) Prepositional Phrase: A feature indicating when an event is part of a prepositional phrase. The feature's values range over 34 English prepositions; 5) Temporal Discourse: I train two models during learning, one for events in the same sentence, and the other for events crossing sentence boundaries. It essentially splits the data on the same sentence feature. SVM is used to classify the event pairs form NYT Corpus with labels 'BEFORE' or 'AFTER' denoting temporal relation between them.

The algorithm further counts the number of times two events are classified as before or after. If the number of one kind of the relation is more than the other over the complete corpus, then the pair is assigned the relation accordingly.

### 3.4 Creating Knowledge Templates

Now I use the unordered event chains and verb pairs with temporal relations to create knowledge templates. Templates are created of the form;

X.**PROP** = *true/false may cause execution of* **A** *[ARG\*: X; ARG\*: Y]*

Intuitively, a statement of the above category means that, the execution of an action A may be triggered if property PROP is true or false for an entity X. Here, A has X as an argument i.e. X participates in the action A. Also, the system annotates the arguments with their role as subject or object. The Fig.2 shows a sample of the extracted knowledge.

TABLE 1
FEATURES USED FOR LEARNING EVENT ATTRIBUTES

| Tense  | POS-2-event, POS-1-event, POS-of-event, have word, be word |
|--------|------------------------------------------------------------|
| Aspect | POS-of-event, modal word, be word                          |
| Class  | Synset                                                     |

## 4 EVALUATIONS

The system was able to use 33.5% of the unordered event chains extracted from the Corpus to create a knowledge base. 5200 unordered event sets were extracted from the corpus.

For qualitative evaluation, I manually filtered the knowledge templates to get 1742 out of 5200 instances that have relevant Event-Based Conditional Commonsense.

## 5 RESULTS

The WSC corpus consists of 282 sentence and question pairs. I focus on a subset of the WSC tasks that requires two specific types of ECC (1) Direct Causal Events – event-event causality and (2) causal attribute [Sharma et al., 2015]. This subcategory contains a total of 71 WSC corpus questions. Out these 71 question the system was able to cor-

```
percent.decline = true may cause execution of fall [percent, production]
percent_object.decline = true may cause execution of fall [percent_object, production_subject]

seats.have = true may cause execution of win [seats, communists]
seats_object.have = true may cause execution of win [seats_object, communists_subject]

group.agree = true may cause execution of accept [deal, group]
group_subject.agree = true may cause execution of accept [deal_object, group_subject]

tommy.sick = true may cause execution of absent [tommy, school]
tommy_subject.sick = true may cause execution of absent [school_object, tommy_subject]
```

Fig. 2. Sample of the extracted knowledge.

rectly answer 22, wrongly answer 8, and did not find relevant knowledge templates for remaining 41 tasks.

The experiment shows that the extracted knowledge from the corpus was useful to tackle 22 questions correctly. The system was not able to attempt 41 questions as it did not have the required knowledge templates for the task. So,

TABLE 2
EXTRACTED KNOWLEDGE STATS

| | |
|---|---|
| Total no. of documents in the corpus | 4011 |
| Total no. of sentences in the corpus | 41771 |
| Avg. no. of sentences per document | 10.40 |
| | |
| Total unordered events extracted | 5200 |
| Total knowledge templates extracted | 1742 |
| Total knowledge templates extracted after filtering for quality knowledge. | 843 |

given a larger corpus the system can extract more quality knowledge.